\documentclass[letterpaper, 10 pt, conference]{ieeeconf}  

\IEEEoverridecommandlockouts                              

\usepackage{caption}
\usepackage[font=footnotesize]{subfig}

\usepackage{cite}
\usepackage{amsmath,amssymb,amsfonts}
\usepackage{algorithmic}
\usepackage{graphicx}
\usepackage{textcomp}
\usepackage{xcolor}
\usepackage{cleveref} 

\usepackage[nolist,nohyperlinks]{acronym}
\newacro{CNN}{Convolutional Neural Network}
\newacro{DNN}{Deep Neural Network}
\newacro{RNN}{Recurrent Neural Network}
\newacro{LSTM}{Long short-term memory}
\newacro{GPS}{Global Positioning System}
\newacro{GNSS}{Global Navigation Satellite System}
\newacro{NLOS}{non-line-of-sight}
\newacro{ADAS}{Advanced Driver Assistance Systems}
\newacro{LIDAR}[LiDAR]{Light Detection And Ranging}
\newacro{HD map}{High Definition map}
\newacro{EV}{Embedding Vector}
\newacro{SLAM}{Simultaneouos Localization And Mapping}
\newacro{MLP}{MultiLayer Perceptron}
\newacro{IMU}{Inertial Measurement Unit}
\newacro{ML}{Machine Learning}
\newacro{SfM}{Structure from Motion}
\newacro{PnP}{Perspective-n-Points}
\newacro{ASPP}{Atrous Spatial Pyramid Pooling}  
\newacro{NHTSA}{National Highway Traffic Safety Administration}
\newacro{BEV}{Bird Eye View}
\newacro{KD}{Knowledge Distillation}
\newacro{MBEV}{Model-Based Bird Eye View}
\newacro{WARPING}{warpings with homography}
\newacro{3DMASKED-BEV}{Masked 3D-generated Bird Eye View}
\newacro{3D-BEV}{3D-generated Bird Eye View}
\newacro{DoF}{Degrees of Freedom}
\newacro{CE}{Cross Entropy loss}
\newacro{FC}{Focal loss}

\def\eg{\emph{e.g., }}
\def\ie{\emph{i.e., }}

\usepackage{booktabs}
\usepackage{adjustbox}
\usepackage{threeparttable}
\usepackage{pifont}
\usepackage{multirow}

\title{\LARGE \bf
Model Guided Road Intersection Classification
}

\author{Augusto Luis Ballardini$^{1}$, Álvaro Hernández Saz and Miguel Ángel Sotelo

\thanks{All authors are from Computer Engineering Department, Polytechnic School, Universidad de Alcalá, Alcalá de Henares, Spain.}
\thanks{$^{1}$His work has been funded by European Union H2020, under GA Marie Sk\l{}odowska-Curie n. 754382 Got Energy.
{\tt\small \{augusto.ballardini, alvaro.hernandezsaz, miguel.sotelo\}@uah.es }}%

}

\usepackage{tikz}
\usepackage{textcomp}
\usepackage{lipsum}

\newcommand\copyrighttext{%
  \footnotesize \textcopyright 2021 IEEE. Personal use of this material is permitted. Permission from IEEE must be obtained for all other uses, in any current or future media, including reprinting/republishing this material for advertising or promotional purposes, creating new collective works, for resale or redistribution to servers or lists, or reuse of any copyrighted component of this work in other works.}
\newcommand\copyrightnotice{%
\begin{tikzpicture}[remember picture,overlay]
\node[anchor=south,yshift=10pt,xshift=10pt] at (current page.south) {\fbox{\parbox{\dimexpr\textwidth-\fboxsep-\fboxrule\relax}{\copyrighttext}}};
\end{tikzpicture}%
}

\newcommand\whichconference{%
  \footnotesize \centering This manuscript has been accepted as a contributed paper to be presented at the 2021 32nd IEEE Intelligent Vehicles Symposium (IV) (IV 2021)\\ July 11-15, 2021, in Nagoya University, Nagoya, Japan.}
\newcommand\conferencenote{%
\begin{tikzpicture}[remember picture,overlay]
\node[anchor=north,yshift=-10pt,xshift=10pt] at (current page.north) {\fbox{\parbox{\dimexpr\textwidth-\fboxsep-\fboxrule\relax}{\whichconference}}};
\end{tikzpicture}%
}

\begin{document}

\maketitle

\copyrightnotice 
\conferencenote

\thispagestyle{empty}
\pagestyle{empty}


\begin{abstract}

Understanding complex scenarios from in-vehicle cameras is essential for safely operating autonomous driving systems in densely populated areas. Among these, intersection areas are one of the most critical as they concentrate a considerable number of traffic accidents and fatalities. Detecting and understanding the scene configuration of these usually crowded areas is then of extreme importance for both autonomous vehicles and modern \ac{ADAS} aimed at preventing road crashes and increasing the safety of vulnerable road users. This work investigates inter-section classification from RGB images using well-consolidate neural network approaches along with a method to enhance the results based on the teacher/student training paradigm. An extensive experimental activity aimed at identifying the best input configuration and evaluating different network parameters on both the well-known KITTI dataset and the new KITTI-360 sequences shows that our method outperforms current state-of-the-art approaches on a per-frame basis and prove the effectiveness of the proposed learning scheme.
\end{abstract}

\label{sec:introduction}
\section{Introduction}

Estimating the scene in front of a vehicle is crucial for safe autonomous vehicle maneuvers and it is also key to advanced \ac{ADAS}. 
Even though over the past years performance and availability of scene understanding systems increased, nowadays technology seems to be far from the requirements of SAE full-automation level, in particular regarding urban areas and contexts without a strict Manhattan-style city planning. 
Among these, intersection areas are one of the most critical, and reports from the United States \ac{NHTSA} show us that intersections concentrate more than 40\% of motor vehicle crashes~\cite{NHTSA}.
Navigation in these areas requires therefore robust systems able to correctly identify them, enabling safe maneuvers as the vehicle approaches and crosses the upcoming intersection.
From an opposite viewpoint, it follows that the detection and moreover the classification of intersection can be used as input to high-level classifiers of drivers maneuvers, or to ease the prediction of position and intentions of vulnerable road users.
Toward this goal, some intersection detectors are tightly coupled with localization procedures that in turn rely on external systems such as \ac{GNSS} or map providers like Google Maps, HERE or TomTom, which started to provide \ac{LIDAR} based maps commonly referred to as \acp{HD map}. 
The benefits of having prior knowledge about the road configuration from maps are undisputed, as allows systems to narrow the localization uncertainty and the plethora of driving scenarios, hence exploit the map data to perform predefined tactical and operational maneuvers.  
However, given the impact of the vehicle crashes, it follows that relying on updated maps might jeopardize the safety of autonomous driving systems themselves. Moreover, \ac{GNSS} reliability in urban areas is frequently hampered by multi-path or \ac{NLOS} issues, requiring for self-sustaining approaches and on-board sensors.
State of the art intersection detection algorithms use a combination of techniques ranging from consolidated computer vision approaches to probabilistic methods to jointly process 3D data from \ac{LIDAR} sensors, images and map features. 
Nevertheless, research progresses during the past years on \acp{DNN} outperformed previous proposals on almost every task, ranging from stereo reconstruction to object detection and image segmentation, according to \cite{lin2014microsoft, deng2009imagenet, cordts2016cityscapes, Menze2015ISA}. 

\begin{figure}[t]
  \begin{center}
  \includegraphics[width=.99\columnwidth]{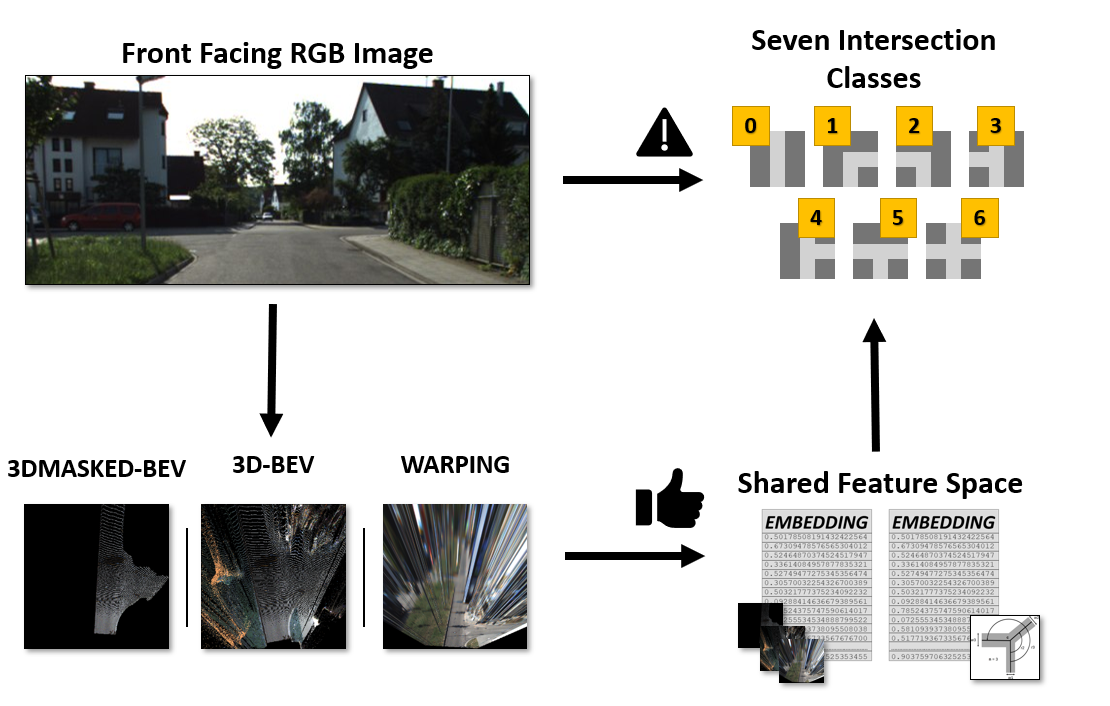}
  \end{center}
  \caption{A short overview of the proposed classification methods.
  Our proposal exploits a synthetic intersection generator to enhance the prediction over standard RGB images, following a teacher/student training scheme.}
  \label{fig:teaser}
  \vspace{-2mm}
\end{figure}

\begin{figure*}[!ht]
  \begin{center}
  \includegraphics[width=1.0\textwidth]{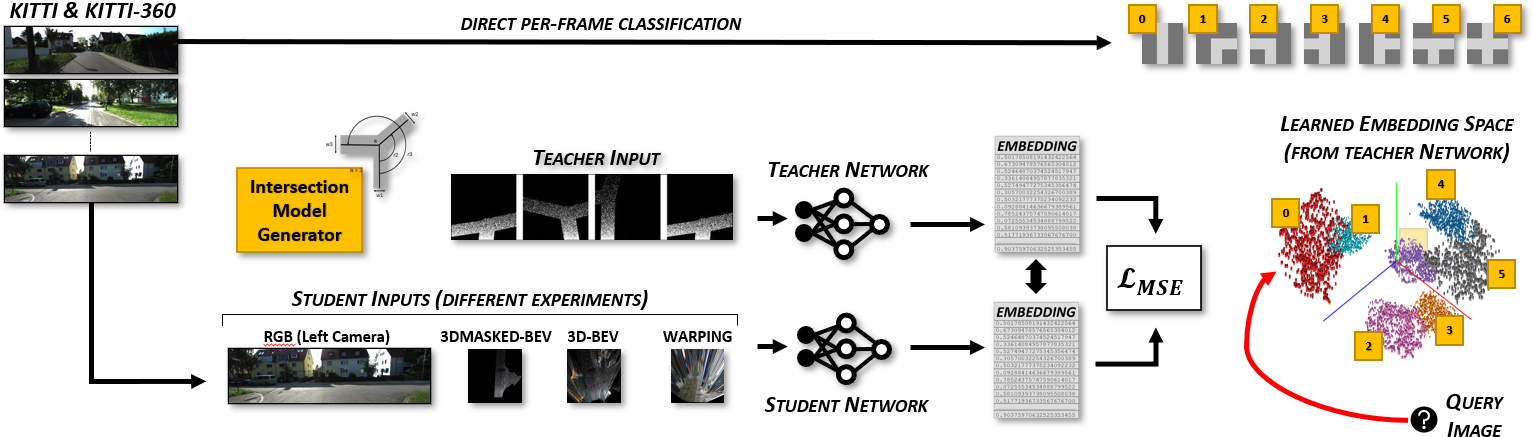}
  \end{center}
  \caption{A schematic description of the performed activities. The upper section depicts the basic classification pipeline by means of well-consolidated \acp{DNN}. In the lower section, the main contribution of this work. It consists of the evaluation of the teacher/student learning paradigm applied to the intersection classification problem, where two \acp{DNN} are trained to obtain similar embedding vectors. The rightmost question mark represents a query using the different inputs (see student inputs).}
  \vspace{-4mm} 
  \label{fig:pipeline}
\end{figure*}

The goal behind this project is to exploit the generalization capabilities of modern \acp{DNN} pushing forward the edges in road intersection classification context. 
Our intent is to understand the typology of intersections in front of a vehicle from RGB sensors only, assessing whether a single-frame technique may serve this purpose and highlighting the limitations. 
This aims at supporting a wide range of advanced driving assistance systems (ADAS) as well as self-driving algorithms, which can greatly benefit from the intersection classification for many sub-task such as localization purposes.
Following our previous works in intersection classification \cite{Ballardini_ICRA_2017, Ballardini_ICRA_2019} and road segmentation \cite{hernandez20203ddeep}, we propose to identify the intersection classes shown in \Cref{fig:teaser}.
Despite the limitation on the seven classes, this allows us to compare the improvements with respect to the previous state of the art, yet paving the way for further investigation on other intersection configurations, \eg roundabouts or large avenues. 
Moreover, differently from previous approaches, our proposal is able to predict the intersection on a frame-per-frame basis without any temporal integration, exceeding the previous state of the art resulting accuracies. 
Specifically we demonstrate that our proposal is effective exploiting both KITTI \cite{Geiger2013IJRR} and new KITTI-360 sequences \cite{Xie2016CVPR}.


\label{sec:related-work}
\section{Related Work}
The relevance of road intersection detection can be noticed from the interest towards the problem coming from different research communities as well as traffic regulation agencies~\cite{Ki2007, golembiewski2011intersection}. 

From a technical perspective, we can first distinguish approaches that exploit images from both stereo or monocular camera-suites, algorithms that only rely on \ac{LIDAR} sensors, or finally a combination of the previous. 
First researches appeared in the intersection detection field date back to the '80s and the works of Kushner and Puri \cite{kushner1987progress}, where matches between road boundaries extracted from images or \ac{LIDAR} sensors data and the correspondent topological maps were exploited to detect intersection areas. 
In different contexts but with a similar approach, authors in \cite{Kim2016, Ballardini_ICRA_2017} exploited RGB images from vehicle front-facing cameras and standard computer vision techniques to create temporally integrated occupancy grids, that were in turn compared to predetermined shapes to assess the presence of upcoming intersections. 
During the same period, the authors in \cite{7727849} proposed a method where three different classifiers were evaluated to distinguish \emph{junctions} from \emph{roads}. Different from the previous method, here 3D pointclouds from \acp{LIDAR} were used.

A second distinction can be done for works involving deep-learning techniques.
Works in this category include the approach in \cite{8206317} with a network called \emph{IntersectNet}, where a sequence of 16 images was passed through an ensemble of \ac{CNN} and \ac{RNN}, combined to set up a three-types intersection classifier (four road-crossing and T-junctions) using a simple average-pooling fusion layer.
A similar ensemble coupled with a more elaborated integration network was used in the work proposed in \cite{Koji2019}. 
Here the authors suggest to use two sets of images relative to the intersection processed with \ac{DNN} and \ac{RNN} respectively.
Regarding the \ac{LIDAR} domain, the authors in \cite{yan2020lidar} proposed a network called LMRoadNet aimed to simultaneously segment road surface and perform topology recognition by an aggregation of consecutive measures. Another interesting approach that exploits \ac{LIDAR} was presented in \cite{8569916}. Here the authors evaluated a transfer-learning method to the classification problem, coupling the sensor readings with the prediction of the ego-vehicle path.

Differently from the previous approaches, we propose a method for classifying upcoming intersections using the \emph{Teacher-Student} learning paradigm, where a combination of two networks is employed. 
The main contribution of this work is then to assess whether this learning method can be used to \emph{guide} the learning of a specific task between different models, usually but not limited to simpler ones. 
In particular, we propose to exploit the simple yet previously assessed intersection model generator presented in \cite{Ballardini_ICRA_2017} as input to the teacher network and then evaluate the learning capabilities of different student network configurations. 

\label{sec:technical-approach}
\section{Technical Approach}
Our intention consists in identifying the topology of the upcoming intersection. 
We used selected sequences from two datasets collected in two different periods in the city of Karlsruhe, Germany \cite{Geiger2013IJRR,Xie2016CVPR}. 
The temporal distance spans over two years, thus we believe it is fair to say that the scenes appear different enough to stress our system generalization capabilities.
As regards the frame selection, we have to make a distinction in the selection process between the two datasets. 
On the one hand, KITTI is distributed with GPS-RTK ground-truth data that allows us to exploit the localization information to automatically select and classify the frames involving intersections, exploiting the cartography of OpenStreetMap. 
Besides speeding up the process, this allowed us to use frames that are up to a specific distance from the intersection center. 
Specifically, we used the selected frames used in \cite{kitti-gt}.
On the other hand, the new KITTI-360 dataset is currently missing the promised OpenStreetMap data as well as per-frame GPS-like positioning. 
This forced us to perform a frame labeling relying only on the appearance of each frame. 
Nevertheless, we were able to manually label all ten sequences of the dataset that were used, alternatively, to train and test the performances of our approach. 
Further details will be provided in \Cref{sec:experimental-results}.

The underlying idea of this work is twofold. 
First, we wanted to prove the capabilities of the teacher/student paradigm in identifying the upcoming intersections, with respect to a basic baseline composed of standard state-of-the-art neural networks. 
Toward this goal, different approaches were experimented and will be described in the following subsections.
Second, this work assesses the classification capabilities of such networks on a frame-by-frame basis, to compare the multi-frame results of previous contributions with the proposed learning paradigm.

An overview of the full pipeline described in this work is proposed in \Cref{fig:pipeline}, and the following subsections explore the extensive experimental activity performed towards our goals.




\begin{figure}[t]
    \begin{center}
        \includegraphics[width=.84\columnwidth]{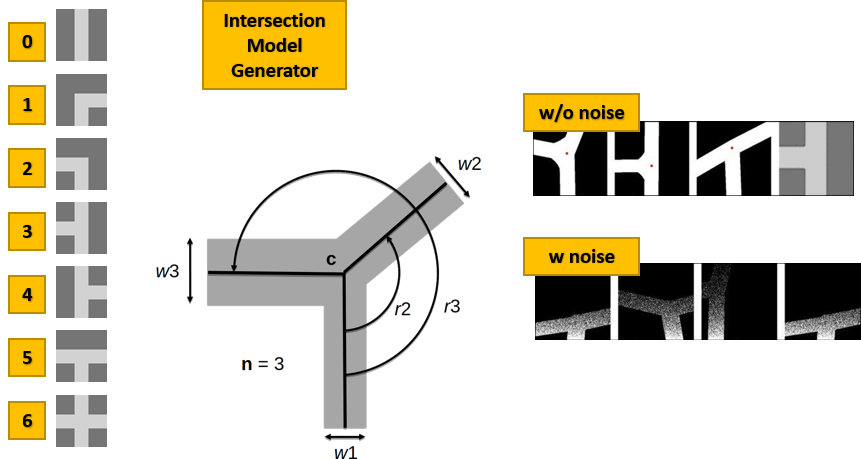}
    \end{center}
    \caption{The seven intersection classes along with the model used to generate the training dataset. 
    In the two following lines: a triplet consisting of two samples of the canonical type-0 (shown in the last box of the row) and a different one, \eg type-5, and an example of the application of the random noise.}
    \label{fig:topology}
\end{figure}

\label{sec:rgb-processing}
\subsection{RGB Pre-processing}
\begin{figure}[t]
  \begin{center}
  \includegraphics[width=.99\columnwidth]{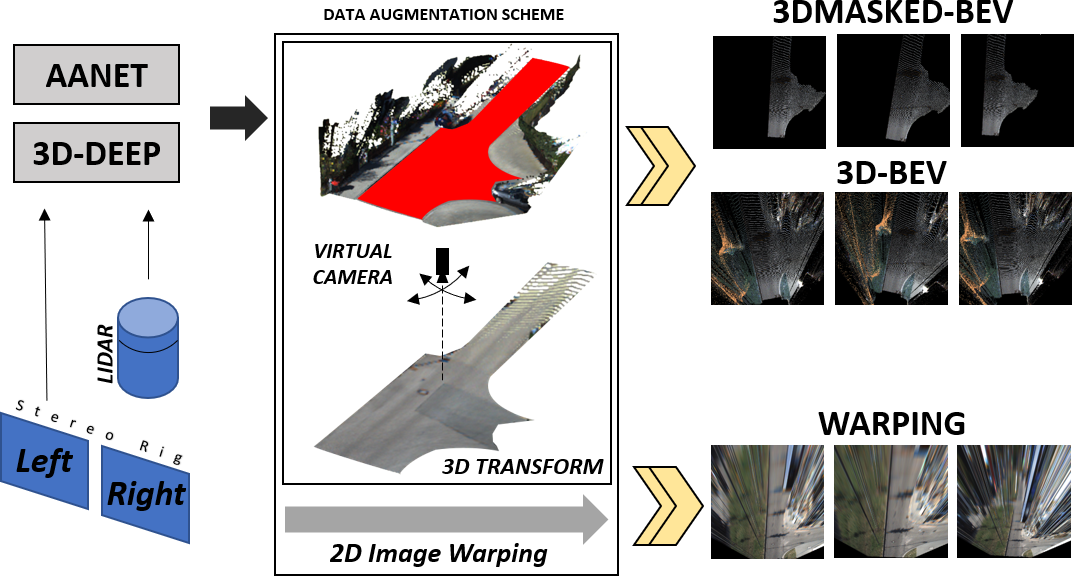}
  \end{center}
  \caption{The figure depicts the pipeline used to generate the images used in this work (apart from the original RGB-left camera). Among them, only the \ac{3DMASKED-BEV} use the \ac{LIDAR} data.}
  \label{fig:RGB-processing}
\end{figure}

To facilitate the comparison with respect to the \acp{MBEV} images generated with the intersection model of \Cref{sec:intersection-model}, we created a pipeline that allows us to transform the RGB images into a similar viewpoint. 
Due to the low amount of frames selected from the first KITTI dataset, we opted for the following scheme, allowing for simultaneous data augmentation process and bird-eye-view image transforms. 
First, using the work in \cite{xu2020aanet} and both the images from the stereo-rig, we created the associated depth-image, which allows us to easily generate a 3D point cloud of the observed scene. We then apply to the depth-image the road segmentation mask obtained using the algorithm presented in \cite{hernandez20203ddeep}, to remove the 3D points that do not belong to the road surface. 
The remaining 3D points are then used to create the so-called \acp{3DMASKED-BEV}, which in turn are very similar to those generated using the intersection model.
Please note that having the 3D coordinates allows us to easily generate as many views as needed even from a single stereo pair, fulfilling the common data-augmentation needs for neural network approaches, see~\Cref{fig:RGB-processing}. 
This allowed us to emulate the sparsity issue of the \acp{3DMASKED-BEV}.
Eventually, to be able to measure the contribution of this additional information to the classification problem, we also generated a version of the images without applying the segmentation mask. We refer to these images as \acp{3D-BEV}. An example of all the possible outcomes resulting from one original single pair of images is shown in \Cref{fig:RGB-processing}.

\label{sec:teacher-student}

\label{sec:intersection-model}
\subsection{Intersection Model}
From a technical perspective and among the possible use cases, the idea behind the teacher/student training paradigm includes transferring knowledge between a more simple domain to a much more complex one. 
In our case, the base domain from which we propose to learn consists of a synthetic set of \ac{BEV} images generated with the intersection model used in the works in \cite{Ballardini_ICRA_2017, Ballardini_ICRA_2019}, for intersection classification and vehicle localization respectively. 
The simple intersection model generator, along with the seven configuration classes, is visually described in \Cref{fig:topology}.
Its complete parameterization includes the possibility to change not only the intersection typology, \eg the number and position of intersecting arms, but also the width of each individual road and the center position with respect to the image.
This model allows us to generate \ac{BEV} binary images containing the shape of all the considered intersections types that can be found in the two datasets, and also an arbitrary amount of them. These will be used during the training phases of our teacher network, acting itself as a data-augmentation scheme for the \ac{DNN}.
We refer to these images as \acp{MBEV}.
At this time, despite its triviality, it should be noted that the point-density of \acp{3DMASKED-BEV} is not constant over the distance with respect to the vehicle. 
Therefore, to simulate comparable \acp{MBEV}, we added a random noise proportional to the distance, see \Cref{fig:topology}. 


\label{sec:baseline}
\subsection{Baseline}
We started our experiments by evaluating the classification capabilities of two well-known network models, namely RESNET-18 and VGG-11 networks, to perform classification in an end-to-end fashion. 
This allowed us to create a first neural-baseline to compare with. 
Please notice that these networks will be then used as backbone for all subsequent activities.
To create this baseline, we first used the RGB images from the left-camera of the stereo rig. 
Alongside, we also prepared a second set of images containing a 2D homographies of the original images, to obtain a so-called \ac{WARPING} images. 
These two sets of images were used in addition to the previous RGB images to perform a comparison between the two representations and then assessing the benefits described in \Cref{sec:related-work}.  
It is worth mentioning that a fair comparison with most existing approaches at this stage is not possible, as they used a frame-integration process. 
Nevertheless, this helped us to set lower-bound thresholds and to evaluate the approaches described in the following subsections.

\subsection{Teacher/Student Training}
In order to compare the images generated from the intersection model and those transformed from the RGB images, we propose a teacher/student paradigm aimed to learn a shared-embedding space between the two domains. 
The approach proposed in this work is inspired by the works of Cattaneo~\cite{Cattaneo2020}, which performs visual localization using 2D and 3D inputs in a bi-directional mode, teaching two networks to create a shared embedding space.
In a similar way, we conceive the classification problem as a metric-learning task where, given two instances of the same intersections class but in different domains, \eg \emph{Class 0} and \emph{Domains D1} and \emph{D2} ($D_{C=0}^{1}$ and $D_{C=0}^{2}$), and two different non-linear functions $f(\cdot)$ and $g(\cdot)$ represented in form of \acp{DNN}, the distance between the embeddings is lower than any other negative intersection instance, \eg $D_{c=2}^{2}$. 
Formally, given the \emph{Intersection-Model} domain $M=\left\{0, 1, ..., 6\right\}$ and \emph{Camera} domain $C=\left\{0, 1, ..., 6\right\}$ each of which contains the seven intersection typologies considered in \Cref{sec:intersection-model}, given one element $m_i \in M$, then \Cref{eq:distance} is satisfied for all the elements in $c_j \in (C \setminus c_i)$, where $d(\cdot)$ is a distance function. 
\begin{equation}
\label{eq:distance}
d(f(m_i), g(c_i)) < d(f(m_i), g(c_{j}))
\end{equation}
With regard to the teacher/student learning scheme, we made the following considerations.
\begin{figure}[t]
  \begin{center}
  \includegraphics[width=.79\columnwidth]{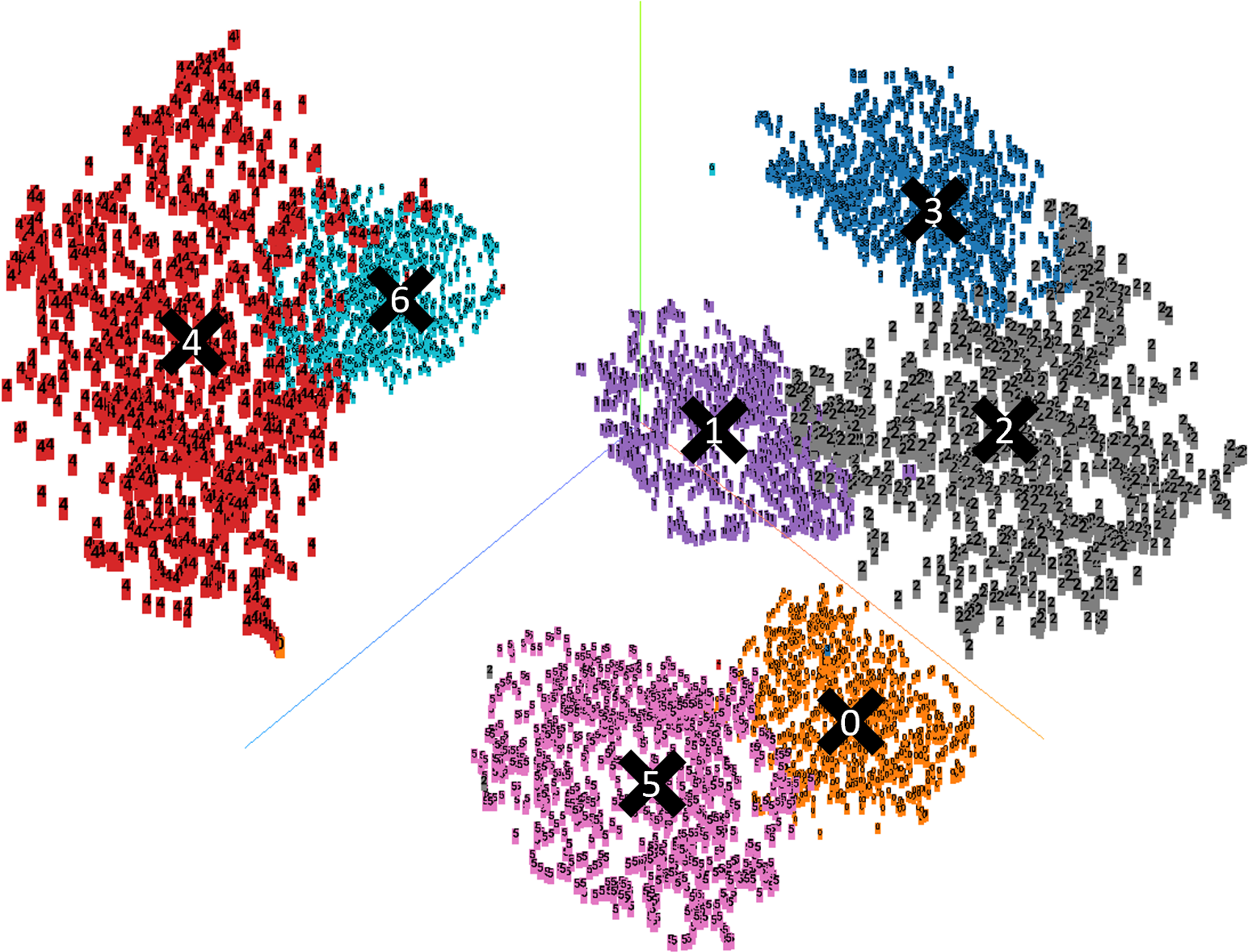}
  \end{center}
  \caption{The embedding space visually represented using T-SNE algorithm.
  In black, we conceptually represent the centroid of each of the clusters.}
  \label{fig:embedding_separation}
\end{figure}
\label{sec:teacher}
\subsubsection{Teacher}
The teacher network is the first of the two networks to be trained. 
It uses the images generated from the intersection model to create a high-dimensional embedding vector associated to each of the seven intersection typologies. 
We used a triplet margin approach~\cite{facenet}, where a set of three images generated with the intersection model $(M_i^a, M_i^s, M_i^d)$ composed of one \emph{anchor} class image $M_i^A$, a \emph{same} class sample $M_i^S$ and a \emph{different} class sample $M_i^D$, is passed through the triplet margin loss function. 
The function is defined similarly to each part of \Cref{eq:distance}, but this time using the same \ac{DNN} model $f(\cdot)$, \ie our teacher network, as follows:
\begin{equation} \label{eq1}
\begin{split}
\mathcal{L}=\sum_i [ & d(f(M_i^A), f(M_i^S)) \\
                   - & d(f(M_i^A), f(M_i^D)) + m]_+
\end{split}
\end{equation}


\noindent
where $[\cdot]_+$ means $max(0, [\cdot])$ and $d(x_i, y_i)= \| x_i - y_i \|_p$ with $p$ as the norm degree for pairwise distance, that in our case was set to $L^2$. As we desire the seven embedding vectors be as much separated as possible, a high separation margin value $m$ was used. \Cref{fig:embedding_separation} depicts the resulting separation.

\begin{table*}[ht!]
    \setlength\tabcolsep{9.6pt} 
    \centering
    \begin{threeparttable}
    \caption{Overall Accuracy Results}
    \label{tab:tableofresults}
        \begin{tabular}{ccccccc|ccc}
        \toprule
        \multicolumn{3}{l}{\multirow{2}{*}{\textbf{Single Sequence Results}}}        & \multicolumn{4}{c}{\textbf{KITTI}}             & \multicolumn{3}{c}{\textbf{KITTI-360}} \\
        \cmidrule(lr){4-7} \cmidrule(lr){8-10} 
        \multicolumn{3}{l}{}                                               & 3DMASKED-BEV (*) & 3D-BEV & WARPING & RGB & 3D-BEV   & WARPING   & RGB   \\ \midrule
        \multirow{4}{*}{Baseline} & \multirow{2}{*}{Resnet18} & C.E. & \ding{55} & \ding{55} & 0.411 & 0.343 & \ding{55} & 0.633 & 0.426 \\
                                  &                           & F.C. & \ding{55} & \ding{55} & 0.467 & 0.345 & \ding{55} & 0.629 & 0.452 \\
                                  & \multirow{2}{*}{VGG11}    & C.E. & \ding{55} & \ding{55} & 0.409 & 0.352 & \ding{55} & 0.729 & 0.562 \\
                                  &                           & F.C. & \ding{55} & \ding{55} & 0.390 & 0.344 & \ding{55} & 0.673 & 0.566 \\ \midrule
        \multirow{2}{*}{Ours}     & \multirow{1}{*}{Resnet18} & MSE   & 0.723 & 0.334 & \textbf{0.514} & \textbf{0.401} & 0.677 & \textbf{0.745} & 0.563 \\ 
                                  & \multirow{1}{*}{VGG11}    & MSE   & 0.687 & 0.221 & 0.381 & 0.315 & 0.602 & \textbf{0.752} & 0.456 \\
        \toprule
        \multicolumn{3}{l}{\multirow{2}{*}{\textbf{Cross Dataset Results}}}        & \multicolumn{4}{c}{\textbf{KITTI}}             & \multicolumn{3}{c}{\textbf{KITTI-360}} \\
        \cmidrule(lr){4-7} \cmidrule(lr){8-10} 
        \multicolumn{3}{l}{}                                               & 3DMASKED-BEV (*) & 3D-BEV & WARPING & RGB & 3D-BEV   & WARPING   & RGB   \\ \midrule
        \multirow{4}{*}{Baseline} & \multirow{2}{*}{Resnet18} & C.E. & \ding{55} & \ding{55} & 0.410 & 0.316 & \ding{55} & 0.599 & 0.597 \\
                                  &                           & F.C. & \ding{55} & \ding{55} & 0.414 & 0.303 & \ding{55} & 0.558 & 0.576 \\
                                  & \multirow{2}{*}{VGG11}    & C.E. & \ding{55} & \ding{55} & 0.417 & 0.327 & \ding{55} & 0.621 & 0.609 \\
                                  &                           & F.L. & \ding{55} & \ding{55} & 0.383 & 0.335 & \ding{55} & 0.625 & 0.625 \\ \midrule
        \multirow{2}{*}{Ours}     & \multirow{1}{*}{Resnet18} & MSE   & 0.723 & 0.315 & \textbf{0.449} & \textbf{0.346} & 0.447 & \textbf{0.640} & 0.615 \\
                                  & \multirow{1}{*}{VGG11}    & MSE   & 0.687 & 0.241 & 0.333 & 0.230 & 0.282 & 0.630 & 0.516 \\
        \bottomrule
        
        \end{tabular}
        \begin{tablenotes}[para, flushleft]
            \footnotesize      
            \item (*) Please note that these results were obtained using the validation set. F.C.: focal-loss. C.E.: Cross Entropy loss.
        \end{tablenotes}
    \end{threeparttable}
    \vspace{-5mm} 
\end{table*}

\subsubsection{Student}
Once the teacher has been trained, we trained the student network using the pre-processed RGB images as input data in a way to obtain a similar embedding vector. 
Towards this goal, the loss-function is composed as follows:
\begin{equation}
\mathcal{L} = \sum_i [d(f(M_i^A), f(C_i^S))]
\label{eq:loss-student}
\end{equation}
where $M$ and $C$ are the model-domain and camera-domain as previously stated and MSE was used as distance function $d(\cdot)$ between the embeddings.
It is worth mentioning that to maintain a consistent distance within same-class classifications, $M_i^A$ elements were chosen not from the list of embedding vectors used in the training phase of the teacher network, but rather from the average of $1000$ new random samples generated after the teacher network was trained, \ie never seen before from the \acp{DNN}. These per-class averages, \ie cluster centroids, are shown in \Cref{fig:embedding_separation} with black crosses, and represent therefore our $M_i^A$ set.

\subsubsection{Training Details}
To avoid overfitting during the training phase of the networks, a data augmentation process was introduced in both networks. 
For what concerns the teacher network, we generated a set of $1000$ per-class intersection configurations by sampling from our generative model. 
We applied a normal random noise to the seven \emph{canonical} intersection configurations on each parameter involved in the generation of the intersection, \eg width, angle and intersection center, in a measure of [$2.0$m, $0.4$rad, $9.0$m]. 
For what concerns the noise, starting from the bottom of the image we added an increasing number of random noise to each line, in a way to mimic the 3D density effect of \acp{3DMASKED-BEV}.
Regarding the student network, since the low number of intersections present in the two KITTI datasets in comparison with the overall number of frames, we performed data augmentation adding a 6-DoF displacement to a looking-down virtual camera originally set at $[10m, 22.5m]$ above the road surface and $[17, 22m]$ in front of the vehicle for the KITTI and \mbox{KITTI-360} respectively. 
Due to the nature of type-1 and type-2 intersection classes, which contains any kind of curve without a specific curvature threshold, we zeroed the rotation along the vertical axis to limit the chance of assimilating these samples to the type-0 class.
Our code leverages the PyTorch 1.6 learning framework~\cite{paszke2017automatic} and both teacher and student images were scaled to images with size 224x224 pixels.

\label{sec:experimental-results}
\section{Experimental Results}

\subsection{Dataset}
To evaluate the classification performances of our approach, we used the following data.
\subsubsection{KITTI}
We used the work in \cite{Ballardini_ICRA_2017} to select \emph{cam\_02|03} color images, raw \ac{LIDAR} readings and GPS-RTK positions of 8 residential sequences, six recorded on 2011/09/30 [18,20,27,28,33,34] and two recorded on 2011/10/03 [27,34]. 
Frames were automatically chosen from the whole sequence by gathering only those that are close up-to $20m$ from the intersection center. We refer the reader to the original publication for further details.
The major issue with this dataset lies with the relatively low number of intersections and the strong imbalance, see \Cref{tab:kitti2011kitti360classes}. 
Considering it would be desirable that all dataset splits, \ie training, validation, and testing, have all types of intersections, the lack of balance forced us to split this dataset only into train and validate splits.  
Please notice that randomly choosing frames from the whole dataset was not an option.
The reason is due to the multiple frames associated with every intersection. 
By randomly selecting frames, it would have been possible to include clearly similar frames of the same intersection into both training and validation or testing, frustrating the separation efforts. 
This left us no choice but to train/validate on this dataset and test on KITTI-360. 
\subsubsection{\mbox{KITTI-360}} 
This dataset contains ten new sequences recorded in 2013, almost two years after the first recordings. Unfortunately, at the moment, no global positioning information is still provided. We then manually labeled the images, including only those images clearly containing intersections, using the images from the previous dataset as a visual-guide. 
This dataset presents a much more balanced set of intersections, allowing us to create good dataset splits and test the previous KITTI dataset.
\begin{figure*}[ht!]
\centering
    \subfloat[Test Seq. - Baseline]{%
    \label{fig:confusion-matrix-01-a}\includegraphics[width=.23\textwidth]{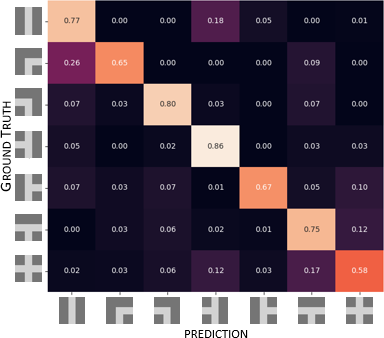}}\quad
    \subfloat[Test Seq. - Ours]{%
    \label{fig:confusion-matrix-01-b}\includegraphics[width=.23\textwidth]{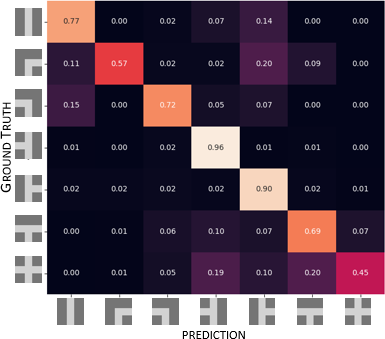}}\quad
    \subfloat[Cross Test - Baseline]{%
    \label{fig:confusion-matrix-01-c}\includegraphics[width=.23\textwidth]{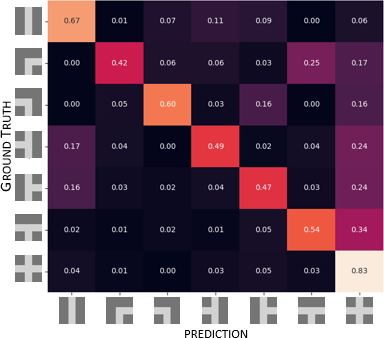}}\quad
    \subfloat[Cross Test - Ours]{%
    \label{fig:confusion-matrix-01-d}\includegraphics[width=.23\textwidth]{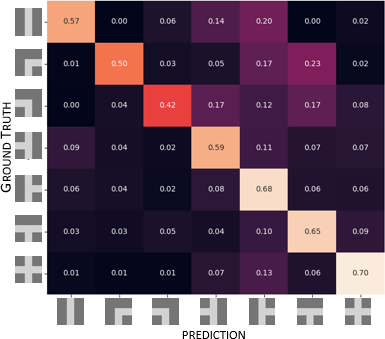}}\quad \\ \vspace{-2mm}
    \subfloat[Test Seq. - Baseline]{%
    \label{fig:confusion-matrix-01-e}\includegraphics[width=.23\textwidth]{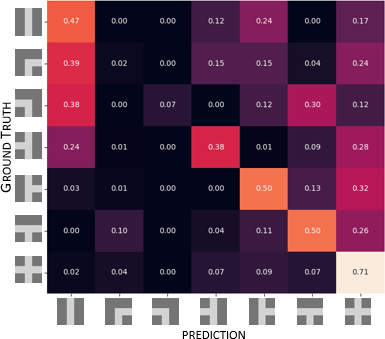}}\quad
    \subfloat[Test Seq. - Ours]{%
    \label{fig:confusion-matrix-01-f}\includegraphics[width=.23\textwidth]{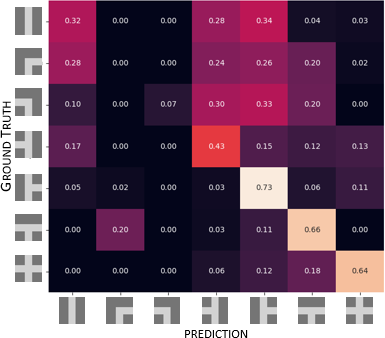}}\quad
    \subfloat[Cross Test - Baseline]{%
    \label{fig:confusion-matrix-01-g}\includegraphics[width=.23\textwidth]{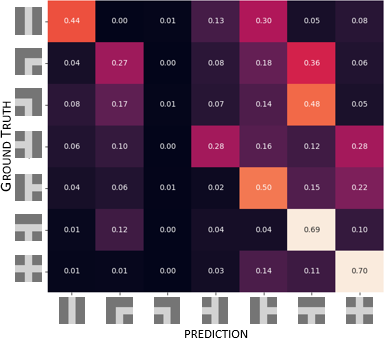}}\quad
    \subfloat[Cross Test - Ours]{%
    \label{fig:confusion-matrix-01-h}\includegraphics[width=.235\textwidth]{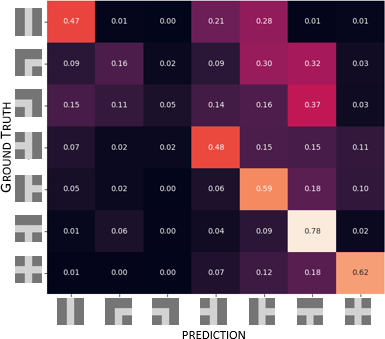}}\quad \\
    \caption{Confusion matrices: first row, training on KITTI-360; second row, training on KITTI. \emph{Test Seq} matrices refer to test executed on the sequence test of same dataset, \emph{Cross Test} refers to experiments with train/test executed on opposite datasets.}
    \label{fig:confusion-matrix-01}
\vspace{-6mm}
\end{figure*}
\begin{table}[hb!]
    \setlength\tabcolsep{3pt} 
    \centering
    \begin{threeparttable}
    \caption{Intersections per-class on evaluated datasets}
    \label{tab:kitti2011kitti360classes}
        \begin{tabular}{crrrrrrr}
        \toprule
                                        \multicolumn{1}{c}{Sequence} & \multicolumn{1}{r}{0} & \multicolumn{1}{r}{1} & \multicolumn{1}{r}{2} & \multicolumn{1}{r}{3} & \multicolumn{1}{r}{4} & \multicolumn{1}{r}{5} & \multicolumn{1}{r}{6} \\ \midrule
        
        2011\_09\_30\_drive\_0018 & 34                    & \ding{55}             & \ding{55}             & 41                    & 23                    & 60                    & 247                   \\          
        2011\_09\_30\_drive\_0020 & 21                    & \ding{55}             & \ding{55}             & \ding{55}             & 45                    & \ding{55}             & 18                    \\
        2011\_09\_30\_drive\_0027 & \ding{55}             & \ding{55}             & \ding{55}             & 25                    & 17                    & 20                    & 152                   \\
        2011\_09\_30\_drive\_0028 & 75                    & 51                    & 19                    & 44                    & 110                   & 131                   & 197                   \\
        2011\_09\_30\_drive\_0033 & 49                    & 39                    & \ding{55}             & 62                    & 17                    & 16                    & 19                    \\
        2011\_09\_30\_drive\_0034 & 15                    & \ding{55}             &\ding{55}              & 77                    & 24                    & 26                    & \ding{55}             \\
        2011\_10\_03\_drive\_0027 & 19                    & \ding{55}             & 25                    & 139                   & 90                    & 183                   & 217                   \\
        2011\_10\_03\_drive\_0034 & 46                    & 49                    & 82                    & 70                    & 113                   & 64                    & 84                    \\\cmidrule(lr){2-8}
        \multicolumn{1}{c}{Total} & 259                   & 139                   & 126                   & 458                   & 439                   & 500                   & 934                   \\
                                        &                       &                       &                       &                       &                       &                       &                       \\[-3mm] \midrule
        2013\_05\_28\_drive\_0000 & 133                   & 46                    & 40                    & 204                   & 153                   & 147                   & 196                   \\
        2013\_05\_28\_drive\_0002 & 4                     & 664                   & 679                   & 200                   & 93                    & 321                   & 93                    \\
        2013\_05\_28\_drive\_0003 & \ding{55}             & 31                    & \ding{55}             & \ding{55}             & \ding{55}             & 27                    & \ding{55}             \\
        2013\_05\_28\_drive\_0004 & 379                   & 154                   & 205                   & 128                   & 125                   & 169                   & 109                   \\
        2013\_05\_28\_drive\_0005 & 69                    & 31                    & 36                    & 76                    & 153                   & 101                   & \ding{55}             \\
        2013\_05\_28\_drive\_0006 & 14                    & 34                    & 66                    & 116                   & 66                    & 132                   & 100                   \\
        2013\_05\_28\_drive\_0007 & 158                   & \ding{55}             & 12                    & 42                    & 11                    & \ding{55}             & 8                     \\
        2013\_05\_28\_drive\_0009 & 318                   & 70                    & 106                   & 305                   & 111                   & 276                   & 622                   \\
        2013\_05\_28\_drive\_0010 & 34                    & 14                    & 59                    & 31                    & 29                    & \ding{55}             & 38                    \\ \cmidrule(lr){2-8}
        \multicolumn{1}{c}{Total} & 1109                  & 1044                  & 1203                  & 1102                  & 741                   & 1173                  & 1166 \\ \bottomrule
        \end{tabular}
        \begin{tablenotes}[para, flushleft]
            \footnotesize      
            \item Frame numbers of KITTI and KITTI-360 respectively. Regarding KITTI, seq. 2011\_09\_30\_drive\_0028 was included in the training set, whilst seq. 2011\_10\_03\_drive\_0034 was used in the validation phase and KITTI-360 was used for testing purposes. 
        \end{tablenotes}
    \end{threeparttable}
\end{table}

\subsection{Evaluation method and Results}
\begin{figure}[b!]
\centering
    \subfloat[Results in \cite{Ballardini_ICRA_2017}]{%
    \label{fig:confusionn-matrix-02-a}
    \includegraphics[width=.40\columnwidth]{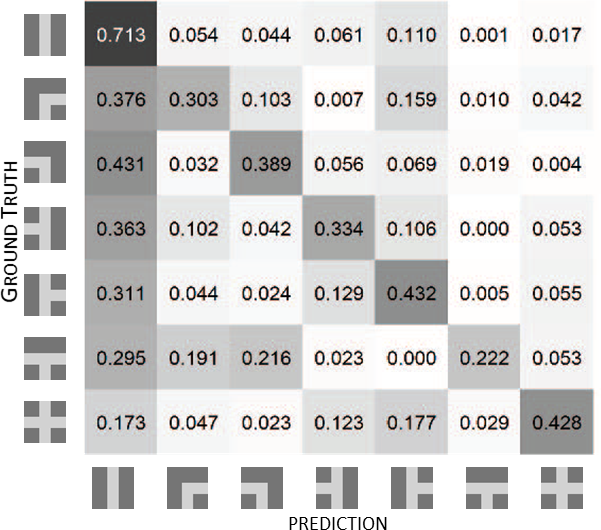}}\quad
    \subfloat[Results in \cite{Koji2019}]{%
    \label{fig:confusionn-matrix-02-b}
    \includegraphics[width=.40\columnwidth]{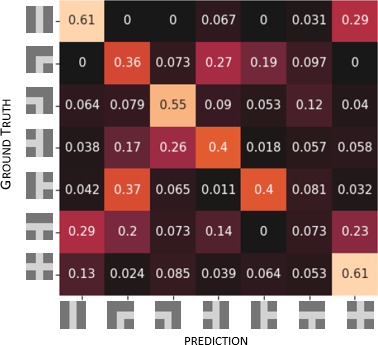}}
    \caption{Confusion matrices of the most similar state of the art classifiers.}
    \label{fig:confusion-matrix-02}
\end{figure}

Regarding the evaluation of the obtained classification, we first created an extensive baseline using the Pytorch implementations of RESNET-18~\cite{He_2016_CVPR} and VGG-11~\cite{simonyan2014very} networks, both with standard \ac{CE} and \ac{FC} \cite{Lin_2017_ICCV} to evaluate different performances. 
From the original KITTI dataset, for each intersection, we select all the frames closer than $20m$ to the intersection center, and selected similar appearing frames from KITTI-360 as no geo-referenced position is still available. 
As opposed to what has been previously done in the works presented in \Cref{sec:related-work}, we performed a per-frame classification aiming at evaluating the abstraction capabilities of modern \acp{DNN}. 
For this reason, the most suitable comparisons concerning the work in \cite{Ballardini_ICRA_2017} are those that consider sequences starting from $20m$ up to the intersection, see \Cref{fig:confusionn-matrix-02-a}. A second comparison can be made with the results of \cite{Koji2019}. In both cases, our approach improved the performances of these contributions.
Results shown in \Cref{tab:tableofresults} and \Cref{fig:confusion-matrix-01} clearly show that our system obtained better results using KITTI-360. 
The lower performances obtained with KITTI might be explained by its strong unbalance, in particular as regards intersection classes 1 and 2. 
Comparing our results to \cite{Ballardini_ICRA_2017}, under the most similar conditions shown in \Cref{fig:confusion-matrix-01-f,fig:confusionn-matrix-02-a}, we can see that our work achieved better results on most classes, except for classes 1 and 2. 
However, the good performances achieved using the KITTI-360 even in cross-dataset experiments support our intuitions on the poor balancing of KITTI sequences, see \Cref{fig:confusion-matrix-01-b,fig:confusion-matrix-01-d}.
The baseline study involved RGB images as well as 2D image-warpings obtained with fixed plane homography. 
Experiments in \Cref{tab:tableofresults} show that the classification obtained using warpings instead of \emph{direct} RGB images obtained better results with both tested backbones, see \Cref{fig:confusion-matrix-01}. 
This led us to state that the classification is better performed using this viewpoint instead of the classic perspective, supporting the model-based learning proposed in this work. 
Finally, even though \ac{FC} outperformed \ac{CE} on RGB images, no significant improvements were achieved using warping images.
Regarding the teacher/student learning paradigm, the experimental activity reported in \Cref{tab:tableofresults}, including testing on both KITTI and KITTI-360 with RESNET-18 and VGG networks and different input data including RGB, \acp{MBEV}, \acp{3DMASKED-BEV}, \acp{3D-BEV} and \acp{WARPING} images, matches or exceed the results of the comparable cases.
\label{sec:conclusions}
\section{Conclusions}
The work presented proposes a comparison of \emph{direct} \ac{DNN}-based intersection classifiers along with an evaluation of the teacher/student training paradigm. 
An extensive experimental activity shows that \ac{DNN} outperforms previous approaches even without any temporal integration.
We demonstrated that the teacher/student allows for a more reliable intersection classification on KITTI datasets.
As the benefits of temporal integration are undisputed, we envision to develop a system to integrate the results of this research as part of our future work. 



\vspace{-2mm}
\bibliographystyle{IEEEtran}
\bibliography{IEEEabrv,model_guided}


\end{document}